\documentclass[a4paper, 12pt]{amia}
\usepackage{graphicx}
\usepackage[labelfont=bf]{caption}
\usepackage[superscript,nomove]{cite}
\usepackage{color}
\usepackage{booktabs}
\usepackage{hyperref}
\usepackage{tabularx,ragged2e,booktabs,caption}
\usepackage[toc,page]{appendix}
\usepackage{mwe}
\usepackage{subcaption}
\usepackage{amsmath}
\usepackage{commath}
\usepackage{listings}
\usepackage{tablefootnote}
\usepackage[utf8]{inputenc}
\usepackage{amsfonts}

\newcolumntype{C}[1]{>{\Centering}m{#1}}

\newcolumntype{L}[1]{>{\raggedleft}m{#1}}

\newcolumntype{R}[1]{>{\raggedright}m{#1}}

\usepackage{sectsty}

\sectionfont{\fontsize{14}{17}\selectfont}

\begin{document}

\title{Adaptively Solving the Local-Minimum Problem for Deep Neural Networks}
\author{Huachuan Wang and James Ting-Ho Lo}

\institutes{
    Department of Mathematics and Statistics, University of Maryland Baltimore County, USA.\\
}

\maketitle

\section{Abstract}
This paper aims to overcome a fundamental problem in the theory and application of deep neural networks (DNNs). We propose a method to solve the local minimum problem in training DNNs directly. Our method is based on the cross-entropy loss criterion's convexification by transforming the cross-entropy loss into a risk averting error (RAE) criterion. To alleviate numerical difficulties, a normalized RAE (NRAE) is employed. The convexity region of the cross-entropy loss expands as its risk sensitivity index (RSI) increases. Making the best use of the convexity region, our method starts training with an extensive RSI, gradually reduces it, and switches to the RAE as soon as the RAE is numerically feasible. After training converges, the resultant deep learning machine is expected to be inside the attraction basin of a global minimum of the cross-entropy loss. Numerical results are provided to show the effectiveness of the proposed method. 

\section{Introduction}
The problem, called the local minimum problem in training DNNs, has plagued the DNN community since the 1980s\cite{2,4}. DNNs trained with backpropagation are extensively utilized to solve various tasks in artificial intelligence fields for decades\cite{3,5,6,7,8}. The computing power of DNNs is derived through its particularly distributed structure and the capability to learn and generalize. However, the application and further development of DNNs have been impeded by the local minimum problem and have attracted much attention for a very long time. DNNs has recently been extensively studied to represent high-level abstractions. Training DNNs involves significantly intricate difficulties, prompting the development of unsupervised layer-wise pre-training and many ingenious heuristic techniques. Although such methods and techniques have produced impressive results in solving famous machine learning tasks, more serious problems originated from the essence of local minimum in high-dimensional non-convex optimization remain\cite{9,10,11,12}. 

A primary difficulty of solving the local minimum problem lies in the intrinsic nonconvexity of the training criteria of the DNNs\cite{19,20,21,22,23}, which usually contain a large number of non-global local minima in the weight space of the DNNs. As the standard optimization methods perform a local search in the parameter (e.g., weight) space, they cannot consistently guarantee the resultant DNN's satisfactory performance even with many training sessions. Although an enormous amount of solutions have been developed to optimize the free parameters of the objective function for consistently achieving a better optimum, these methods or algorithms cannot solve the local minimum problem essentially with the intricate presence of the non-convex function\cite{13,14,15,16,17,27}.

The most standard
approach to optimize DNNs is Stochastic Gradient
Descent (SGD). There are many variants of SGD, and researchers
and practitioners typically choose a particular variant
empirically. While nearly all DNNs optimization algorithms
in popular use are gradient-based, recent work has
shown that more advanced second-order methods such as
L-BFGS and Saddle-Free Newton (SFN) approaches can
yield better results for DNN tasks \cite{25}\cite{22}. Second-order derivatives can be addressed
by GPUs or batch
methods when dealing with massive data, SGD still provides
a robust default choice for optimizing DNNs.
Instead of modifying the network structure or optimization
techniques for DNNs, we focused on designing a new
error function to convexify the error space. The convexification
approach has been studied in the optimization community
for decades but has never been seriously applied within
deep learning. A well-known methods is the 
LiuFloudas convexification method\cite{2, 31}.
LiuFloudas convexification can be applied to optimization
problems where the error criterion is twice continuously differentiable,
although determining the weight $\alpha$ of the added
quadratic function for convexifying the error criterion involves
significant computation when dealing with massive
data and parameters.
Following the same name employed for deriving robust
controllers and filters\cite{29,30,31}.

To alleviate the local minimum problem's fundamental difficulty in training DNNs, this paper proposes a series of methodologies by applying convexification and deconvexification to avoid non-global local minima and achieve the global or near-global minima with satisfactory optimization and generalization performances. These methodologies are developed based on a normalized risk-averting error (NRAE) criterion. The use of this criterion removes the practical difficulty of computational overflow and ill-initialization that existed in the risk-averting error criterion, which was the predecessor of the NRAE criterion Furthermore, it has benefits to effectively handle non-global local minima by convexifying the non-convex error space.
The method's effectiveness based on the NRAE criterion is evaluated in training multilayer perceptrons (MLPs) for function approximation tasks, demonstrating the optimization
advantage compared to training with the standard mean squared error criterion. Moreover, numerical experiments also illustrate that the NRAE-based training methods applied to train DNNs, such as convolutional neural networks and deep MLPs, to recognize handwritten digits in the MNIST dataset achieve better optimization generalization results than many benchmark performances. Finally, to enhance the generalization of the DNNS obtained with the NRAE-based training, a statistical pruning method that prunes redundant connections of the DNNS is implemented and confirmed for further improving the generalization ability of the DNNs trained by the NRAE criterion.

\section{Risk-Averting Error}
Given training samples ${X, y} = {(x_1, y_1), (x_2, y_2), ..., (x_m, y_m)}$, the function $f(x_i,W)$
is the learning model with parameters W. The cross-entropy loss function $J(\boldsymbol{W})$ is defined as:

\begin{equation}
J\left(f\left(\boldsymbol{x}_{i}, \boldsymbol{W}\right), y_{i}\right)=\frac{1}{m} \sum_{i=1}^{m}\left[y_{i}\log f\left(\boldsymbol{x}_{i}, \boldsymbol{W}\right)+(1-y_{i})\log \left(1-f(\boldsymbol{x}_{i}, \boldsymbol{W})\right)\right]
\end{equation}

The Risk-Averting Error criterion (RAE) corresponding to the $L(\boldsymbol{W})$ is defined by

\begin{equation}
R A E_{p}\left(f\left(\boldsymbol{x}_{i}, \boldsymbol{W}\right), y_{i}\right)=\frac{1}{m} \sum_{i=1}^{m} e^{\lambda^{p}\left[y_{i}\log f\left(\boldsymbol{x}_{i}, \boldsymbol{W}\right)+(1-y_{i})\log \left(1-f(\boldsymbol{x}_{i}, \boldsymbol{W})\right)\right]}
\end{equation}

\begin{equation}
=\frac{1}{m} \sum_{i=1}^{m} \left[ f\left(\boldsymbol{x}_{i}, \boldsymbol{W}\right)^{y_{i}}\left(1-f(\boldsymbol{x}_{i}, \boldsymbol{W})\right)^{(1-y_{i})} \right] ^{\lambda^{p}}
\end{equation}

$\lambda$ is the convexity index. It controls the size of the convexity
region. 

Because RAE has the sum-exponential form, its Hessian matrix is tuned exactly by the convexity index $\lambda^{p}$. Given the Risk-Averting Error criterion $RAE_{p} (p  \in \mathcal{N}^{+}$, which is twice continuous
differentiable. $J_{p}(W)$ and $H_{p}(W)$ are the corresponding
Jacobian and Hessian matrix. As $\lambda  \to \infty $,
the convexity region monotonically expands to the entire
parameter space except for the subregion $S := \{W \in
R^n | rank(H_{p}(W)) < n, H_{p}(W < 0)\}$.

Intuitively, the use of the RAE was motivated by its emphasizing
large individual deviations in approximating functions
and exponentially optimizing parameters,
thereby avoiding such large individual deviations and
achieving robust performances. When the convexity index $\lambda$ increases to infinity,
the convexity region in the parameter space of RAE expands
monotonically to the entire space except for the intersection of a
the finite number of lower-dimensional sets. The number of sets
increases rapidly as the number m of training samples increases.
Roughly speaking, larger $\lambda$ and m cause the size of
the convexity region to grow larger respectively in the error
space of RAE\cite{1}.

When $\lambda \to \infty$, the error space can be perfectly stretched
to be strictly convex, thus avoid the local optimum to guarantee
a global optimum. Although RAE works well in theory,
it is not bounded and suffers from the exponential magnitude
and arithmetic overflow when using gradient descent in
implementations.

\section{Normalized Risk-Averting Error}

\begin{equation}
\begin{split}
&N R A E_{p}\left(f\left(\boldsymbol{x}_{i}, \boldsymbol{W}\right), y_{i}\right) \\
&=\frac{1}{\lambda^{p}} \log R A E_{p}\left(f\left(\boldsymbol{x}_{i}, \boldsymbol{W}\right), y_{i}\right) \\
&=\frac{1}{\lambda^{p}} \log \frac{1}{m}  \sum_{i=1}^{m} \left[ f\left(\boldsymbol{x}_{i}, \boldsymbol{W}\right)^{y_{i}}\left(1-f(\boldsymbol{x}_{i}, \boldsymbol{W})\right)^{(1-y_{i})} \right] ^{\lambda^{p}}
\end{split}
\end{equation}
If $RAE_{p}(f(x_i,W), y_i)$ is convex, it is quasiconvex.
$\log$ function is monotonically increasing, so the
composition $\log{RAE_{p}(f(x_i,W), y_i)}$ is quasi-convex. $\log$ is a strictly monotone function and
$NRAE_{p}(f(x_i,W), y_i)$ is quasi-convex, so it shares the same local and global minimizer with $RAE_{p}(f(x_i,W), y_i)$\cite{1}.

The convexity region of NRAE is consistent with RAE.
To interpret this statement in another perspective, the log
function is a strictly monotone function. Even if RAE is
not strictly convex, NRAE still shares the same local and
global optimum with RAE. If we define the mapping function
$f : RAE  \to NRAE$, it is easy to see that f is bijective
and continuous. Its inverse map $f^{-1}$ is also continuous, so
that f is an open mapping. Thus, it is easy to prove that the
mapping function $f$ is a homeomorphism to preserve all the
topological properties of the given space.
The above theorems state the consistent relations among
NRAE, RAE and cross-entropy loss. It is proven that the greater the convexity
index $\lambda$, the larger is the convex region is. Intuitively,
increasing $\lambda$ creates tunnels for a local-search minimization
procedure to travel through to a good local optimum.
However, we care about the justification on the advantage
of NRAE. 

Given training samples $\left\{\boldsymbol{X}, y\right\} = \left\{(\boldsymbol{x_1}, y_1),(\boldsymbol{x_2}, y_2), ...,(\boldsymbol{x_m}, y_m)\right\}$ and the model $f(x_i, \boldsymbol{W})$ with parameters $W$.  If $\lambda^p \geq 1, p \in
\mathcal{N}^{+}$, then both $RAE_{p}(f(x_i,W), y_i)$ and
$NRAE_{p}(f(x_i,W), y_i)$ always have the higher chance to
find a better local optimum than the cross-entropy error
due to the expansion of the convexity region.

Because
$NRAE_{p}(f(x_i,W), y_i)$ is quasi-convex, sharing the same
local and global optimum with $RAE_{p}(f(x_i,W), y_i)$, the
above conclusions are still valid.

Roughly speaking, NRAE always has a larger convexity
region than the  cross-entropy error in terms of their
Hessian matrix when $\lambda \geq 1$. This property guarantees the
higher probability to escape poor local optima using NRAE.
In the worst case, NRAE will perform as good as standard
cross-entropy error if the convexity region shrinks as $\lambda$ decreases
or the local search deviates from the ?tunnel? of convex regions.
More specifically, $NRAE_{p}(f(x_i,W), y_i)$ approaches the standard Lp-norm error as $\lambda_p \to 0$ and approaches the minimax error criterion in $f_W \alpha_{max}(W)$
as $\lambda_p \to \infty$.

\section{Learning Methods}
Under some regularity conditions, the convexity region of $J(\boldsymbol{W})$ expands
monotonically as $\lambda$ increases. To make advantage of a larger convexity region of  $J(\boldsymbol{W})$ at a greater $\lambda$ and avoid computer overflow, we are tempted to minimize $N R A E_{p}\left(f\left(\boldsymbol{x}_{i}, \boldsymbol{W}\right), y_{i}\right)$
at a $\lambda$
as large as possible. However, at a very large $\lambda$, the training process is extremely slow
or grinds to a halt, which phenomenon is called training stagnancy:

minimizing $f_W \alpha_{max}(W)$ for $\lambda \gg 1$ minimizes virtually the
largest $f_W \alpha_{max}(W)$. The architecture of
the deep learning machine is therefore redundant for the approximation. When all the weights are adjusted
to achieve the approximation, they tend to become similar or duplicated, thus
causing rank deficiency, violating the regularity conditions required for convexification
of  $J(\boldsymbol{W})$.

In the Adaptive Normalized Risk-Avering Training
(ANRAT) approach\cite{1,4,12}, we learn $\lambda$ adaptively in
error backpropagation by considering $\lambda$ as a parameter instead
of a hyperparameter. The learning procedure is standard
batch SGD. We show it works quite well in theory and
practice.
The loss function of ANRAT is
\begin{equation}
l(\boldsymbol{W}, \lambda) =\frac{1}{\lambda^{p}} \log \frac{1}{m}  \sum_{i=1}^{m} \left[ f\left(\boldsymbol{x}_{i}, \boldsymbol{W}\right)^{y_{i}}\left(1-f(\boldsymbol{x}_{i}, \boldsymbol{W})\right)^{(1-y_{i})} \right] ^{\lambda^{p}} + a \| \lambda \|^{-q}
\end{equation}

We also use a penalty term a $\| \lambda \|^{-q}$
to control the changing rate of $\lambda$. While minimize the NRAE
score, small $\lambda$ is penalized to regulate the convexity region.
a is a hyperparameter to control the penalty index. $ \alpha_{i}(\mathbf{W}) = f(\mathbf{x}_i, \mathbf{W})^{y_i\lambda^p-1}$, $ \beta_{i}(\mathbf{W}) =(1- f(\mathbf{x}_i, \mathbf{W}))^{\lambda^p-\lambda^py_i - 1}$ and $\gamma_{i}(\mathbf{W}) = f\left(\boldsymbol{x}_{i}, \boldsymbol{W}\right)^{y_{i}}\left(1-f(\boldsymbol{x}_{i}, \boldsymbol{W})\right)^{(1-y_{i})} $. The first order
derivatives on weight and $\lambda$ are
\begin{equation}
\begin{aligned}
\frac{d l(W, \lambda)}{d W} &=\frac{ \sum_{i=1}^{m} \alpha_{i}(\mathbf{W}) \beta_{i}(\mathbf{W})f(\mathbf{x}_i, \mathbf{W})(1-f(\mathbf{x}_i, \mathbf{W}))\lambda^p(y_i - f(\mathbf{x}_i, \mathbf{W}))\frac{\partial f\left(\boldsymbol{x}_{i}, \boldsymbol{W}\right)}{\partial \boldsymbol{W}}}{\sum_{i=1}^{m} \alpha_{i}(\mathbf{W}) \beta_{i}(\mathbf{W}) } \\
\frac{d l(W, \lambda)}{d \lambda} &=\frac{-p}{\lambda^{p+1}}  \log \frac{1}{m}  \sum_{i=1}^{m} \left[ f\left(\boldsymbol{x}_{i}, \boldsymbol{W}\right)^{y_{i}}\left(1-f(\boldsymbol{x}_{i}, \boldsymbol{W})\right)^{(1-y_{i})} \right] ^{\lambda^{p}} \\
&+\frac{1}{\lambda^p} \frac{\sum_{i=1}^{m} p\lambda^{p-1} \log \gamma_{i}(\mathbf{W})\gamma_{i}(\mathbf{W})^{\lambda^p}}{\sum_{i=1}^{m}\gamma_{i}(\mathbf{W})^{\lambda^p}}\\
&-\quad a q \lambda^{-q-1}
\end{aligned}
\end{equation}

\begin{equation}
\begin{aligned}
\frac{d l(W, \lambda)}{d \lambda} 
& \approx \frac{q}{\lambda}\left(L-\text { cross entropy error }-N R A E\right)
\end{aligned}
\end{equation}

This training approach has more flexibility. The gradient
on $\lambda$ as the weighted difference between NRAE and
the standard cross entropy error, enables NRAE to approach the
cross entropy error by adjusting $\lambda$ gradually. Intuitively, it keeps
searching the error space near the manifold of the cross entropy
error to find better optima in a way of competing with and at
the same time relying on the standard cross entropy error space.
The penalty weight a and index q control the
convergence speed by penalizing small $\lambda$. Smaller a emphasizes
tuning $\lambda$ to allow faster convergence speed between
NRAE and cross entropy error. Larger a forces larger $\lambda$ for a
better chance to find a better local optimum but runs the risk
of plateaus and deviating far from the stable error space. q
regulates the magnitude of $\lambda$ and its derivatives in gradient
descent.

This loss function is minimized by batch SGD without
complex methods, such as momentum, adaptive/hand tuned
learning rates or tangent prop. The learning rate and penalty
weight a are selected in $\{1, 0.5, 0.1\}$ and $\{1, 0.1, 0.001\}$
on validation sets respectively. The initial $\lambda$ is fixed at 10.
We use the hold-out validation set to select the best model,
which is used to make predictions on the test set. All experiments
are implemented quite easily in Python and Theano
to obtain GPU acceleration\cite{27}.
The MNIST dataset\cite{32} consists of hand
written digits $0-9$ which are $28\times28$ in size. There are $60,000$
training images and $10,000$ testing images in total. We use
10000 images in training set for validation to select the hyperparameters
and report the performance on the test set.We
test our method on this dataset without data augmentation.
\section{Results and Discussion}
On the MNIST dataset we use the same structure of LeNet5
with two convolutional max-pooling layers but followed by
only one fully connected layer and a densely connected softmax
layer. The first convolutional layer has 20 feature maps
of size $5\times5$ and max-pooled by $2\times2$ non-overlapping windows.
The second convolutional layer has $50$ feature maps.
with the same convolutional and max-pooling size. The fully
connected layer has $500$ hidden units. An $l2$ prior was used
with the strength $0.05$ in the Softmax layer. Trained by ANRAT,
we can obtain a test set error of $0.56\%$, which is the
best result we are aware of that does not use dropout on the
pure ConvNets. We summarize the best published results on
the standard MNIST dataset in Table 1.

\begin{center}
\begin{tabular}{lll}
\hline
\hline
 &Method &Error $\%$    \\
\hline
&ConvNets + ANRAT  &0.56 \\
&ConvNets + ANRAT + dropout  &0.33 \\\hline
\end{tabular}
\captionof{table}{Test set misclassification rates of the best methods
that utilized convolutional networks on the original MNIST
dataset using single model.} 
\end{center}

\section{Conclusion}
Six advantages of the proposed method:
\begin{itemize}
\item no need for repeated trainings (or consistent performances among different training sessions with different initialization seeds); 
\item applicability to virtually any data fitting; 
\item conceptual simplicity and math-ematical justification; 
\item possibility of its use jointly with other training methods; 
\item a smaller DLM with the same or similar performance; 
\item a same-architecture DLM with a better performance.
\end{itemize}

\newpage

\bibliographystyle{unsrt}
\bibliography{main}

\end{document}